\documentclass[letterpaper, 10 pt, conference]{ieeeconf}  

\IEEEoverridecommandlockouts                              

\usepackage{graphics} 
\usepackage{epsfig} 
\usepackage{mathptmx} 
\usepackage{amsmath} 
\usepackage{amssymb}  
\usepackage{multirow}
\usepackage{url}
\usepackage[table]{xcolor}

\usepackage{newtxtext, newtxmath} 

\usepackage{algorithm}
\usepackage{algpseudocode}

\algnewcommand\algorithmicinput{\textbf{Input:}}
\algnewcommand\INPUT{\item[\algorithmicinput]}
\algnewcommand\algorithmicoutput{\textbf{Output:}}
\algnewcommand\OUTPUT{\item[\algorithmicoutput]}

\title{\LARGE \bf
Robot Policy Learning from Demonstration Using Advantage Weighting and Early Termination
}

\author{Abdalkarim Mohtasib$^{1}$, Gerhard Neumann$^{2}$, Heriberto Cuay\'ahuitl$^{1}$
\thanks{$^{1}$Lincoln Center for Autonomous Systems (L-CAS), University of Lincoln, Lincoln, United Kingdom.
        {\tt\small amohtasib@lincoln.ac.uk}}%
\thanks{$^{2}$Autonomous Learning Robots, KIT, Karlsruhe, Germany}
}

\begin{document}

\maketitle
\thispagestyle{empty}
\pagestyle{empty}

\begin{abstract}

Learning robotic tasks in the real world is still highly challenging and effective practical solutions remain to be found. 
Traditional methods used in this area are imitation learning and reinforcement learning, but they both have limitations when applied to real robots. Combining reinforcement learning with pre-collected demonstrations is a promising approach that can help in learning control policies to solve robotic tasks. In this paper, we propose an algorithm that uses novel techniques to leverage offline expert data using offline and online training to obtain faster convergence and improved performance. The proposed algorithm (AWET) weights the critic losses with a novel agent advantage weight to improve over the expert data. In addition, AWET makes use of an automatic early termination technique to stop and discard policy rollouts that are not similar to expert trajectories---to prevent drifting far from the expert data. In an ablation study, AWET showed improved and promising performance when compared to state-of-the-art baselines on four standard robotic tasks. 

\end{abstract}

\section{INTRODUCTION}

Reinforcement Learning (RL) has seen significant success in sequential decision-making domains such as game playing \cite{mnih2015human,silver2016mastering}. But its application to robotics is still challenging due to the exploration problem and costly collection of data. Standard RL algorithms are data-inefficient and require thousands if not millions of interactions with the environment to learn good policies \cite{lillicrap2016continuous,fujimoto2018addressing,haarnoja2018soft,schulman2017proximal}. While it is possible to collect offline expert data for solving robotic tasks, applying deep RL algorithms (though useful to  train large models) using  pre-collected offline data from the real-world is still conceptually and practically  challenging. 
Decision-making control problems for robotic tasks are usually solved using RL-based imitation learning. Mimicking expert behaviours using imitation learning methods has seen some successes  \cite{zhang2018deep,mandlekar2020learning,ebert2021bridge,mandlekar2022matters,kalakrishnan2009learning,nakanishi2004learning,giusti2015machine,pomerleau1989alvinn,bojarski2016end}. Yet, they suffer from distribution shift problems \cite{ross2011reduction,levine2020offline}. Similarly, pure RL methods have also seen some successes in robotics \cite{lillicrap2015continuous,pinto2016supersizing,levine2018learning,gu2017deep,levine2016end,johannink2019residual}. However, most of these results have been achieved in simulation due to the difficulty of applying RL to real robots.

The most successful results in this area have been achieved by combining RL with imitation learning \cite{piot2014boosted,hester2018deep,vecerik2017leveraging,zuo2020efficient,kang2018policy,Rajeswaran-RSS-18,nair2020accelerating,wang2020critic,peng2018deepmimic}. Reinforcement learning from Demonstrations (RLfD) methods use offline collected expert data to bootstrap policies, which helps to mitigate the exploration problem and speeds up the policy convergence in the online learning stage. Many algorithms have been proposed in this area such as  \cite{hester2018deep,vecerik2017leveraging,nair2018overcoming,kang2018policy,Rajeswaran-RSS-18,nair2020accelerating,wang2020critic,peng2018deepmimic,gao2018reinforcement}. Yet, RLfD algorithms are still not well developed, and there is still room for improvement. 

\begin{figure}[t!]
    \centerline{\includegraphics[width=0.85\columnwidth]{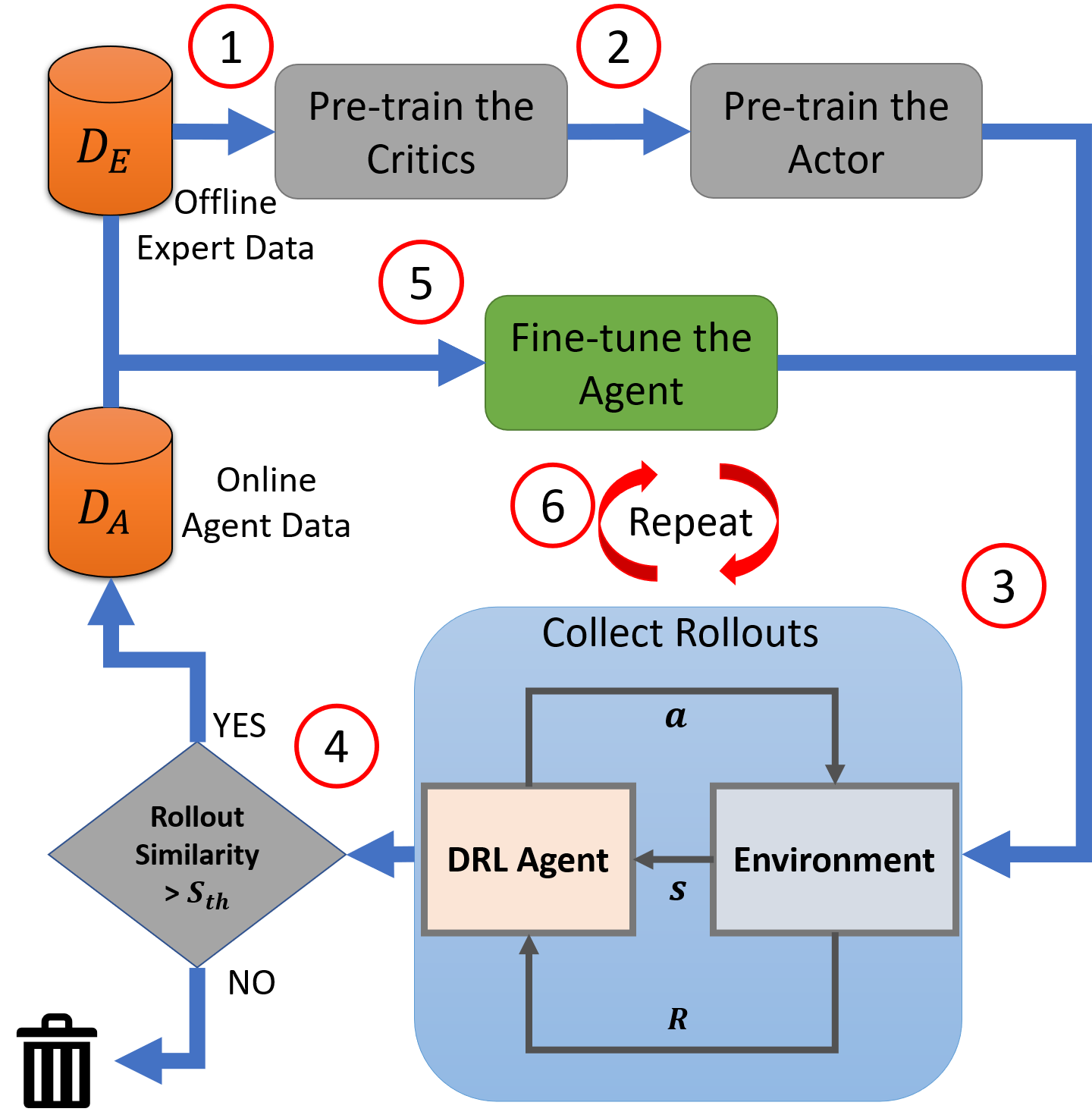}}
    \caption{Overview of our proposed method: (1) pre-train the critics using the expert data; (2) pre-train the actor using expert data; (3) collect policy rollouts that are similar to the expert trajectories; (4) check the similarity of the rollout to expert data; (5) fine-tune the agent using both the expert and agent data; and (6) repeat until convergence.}
    \label{fig_overview}
\end{figure}

In this paper, we propose an advantage weighting with early termination actor-critic (AWET) algorithm that uses offline data to pre-train the agent, followed by  fine-tuning the  agent using both  offline data and agent-environment interactions, see Fig.~\ref{fig_overview}. AWET uses two novel components. The first is weighting the critic losses with a novel agent advantage weight. The second is an automatic early termination technique that discards any policy rollout that is not similar to  expert data. AWET has been compared against  state-of-the-art RL and RLfD algorithms and showed superior performance. 
This is supported by an ablation study of the different components of AWET. The code, models, videos, and data used in this work will be made publicly available\footnote{\url{https://Mohtasib.github.io/AWET_RL/}}.

\section{RELATED WORK}


Typical learning methods for solving decision-making problems in robotics include imitation learning and reinforcement learning. Imitation learning techniques (also called learning from demonstrations \cite{schaal1997learning}) aim to mimic human behaviour in a given task \cite{hussein2017imitation}. An agent is trained to perform a task from demonstrations by learning a mapping between observations and actions \cite{hussein2017imitation}. In reinforcement learning (RL), an agent tries to maximise the expected future rewards by interacting with the environment \cite{sutton2018reinforcement}. Our algorithm combines these two paradigms to solve robotic tasks.

{\bf Imitation Learning}. It is a classical technique that aims at mimicking expert behaviours \cite{schaal1997learning}. Perhaps, the most common imitation learning approach is behavioural cloning (BC), which uses supervised learning methods to learn a policy from expert state-action pairs \cite{schaal1997learning,hussein2017imitation}. BC has achieved some successful results in robot manipulation \cite{zhang2018deep,mandlekar2020learning,ebert2021bridge,mandlekar2022matters}, locomotion \cite{kalakrishnan2009learning,nakanishi2004learning}, navigation \cite{giusti2015machine}, and autonomous driving \cite{pomerleau1989alvinn,bojarski2016end}. Dataset Aggregation (DAGGER) \cite{ross2011reduction}, a popular imitation learning algorithm, augments a dataset by interleaving learned and expert policies to eliminate the accumulated error. Sun et al. \cite{sun2017deeply} developed the Deeply AggreVaTeD algorithm as an extended version of DAGGER that works with deep neural nets and continuous action spaces. However, they both suffer from the same problem, the need for the expert demonstrator to provide actions during the whole training steps, which makes them difficult to use in practice \cite{ross2011reduction,sun2017deeply}. Furthermore, Generative Adversarial Imitation Learning (GAIL) \cite{ho2016generative} is an imitation learning algorithm that has shown effectiveness in high-dimensional continuous control problems. It uses a discriminator to distinguish whether state-action pairs are from the expert or the learned policy \cite{ho2016generative}. Meanwhile, GAIL optimises the policy to confuse the discriminator \cite{ho2016generative}. While it is sample efficient with expert data, it is not during online learning.

Fundamentally, pure imitation learning approaches are limited because they suffer from sharp performance decline when the expert data is imperfect \cite{hussein2017imitation,ross2011reduction}. They also cannot outperform the demonstrator since they lack a notion of task performance \cite{levine2020offline}. Additionally, they suffer from distribution shift problems \cite{ross2011reduction,levine2020offline}. The most recent successful imitation learning algorithms are based on Inverse
Reinforcement Learning (IRL) \cite{ng2000algorithms}, where a reward function is inferred from the demonstrations without knowing the dynamics \cite{mohtasib2021neural}. IRL has seen success in many tasks such as manipulation \cite{finn2016guided}, autonomous helicopter flight \cite{abbeel2004apprenticeship}, and navigation \cite{ziebart2008maximum}. However, in our work, we assume knowledge of the reward function. So, we omit comparisons to IRL methods.

{\bf Reinforcement Learning} (RL). Although RL approaches are hard to apply to robotics tasks, they have been extensively investigated because of the autonomy they could achieve. RL-based robots have been able to balance a unicycle \cite{deisenroth2011pilco}, swing up a cart-pole, and play table tennis \cite{peters2010relative}. The interest in applying RL to robotics has increased after the success of RL in video/board games like Atari Games \cite{mnih2015human} and Go \cite{silver2016mastering}. Also, combining RL algorithms with large deep neural networks led to Deep Reinforcement Learning (DRL) which helped to learn  policies from high-dimensional observations (i.e. raw pixels). DRL has achieved significant results when applied to robotic tasks such as locomotion \cite{lillicrap2015continuous}, grasping \cite{pinto2016supersizing,levine2018learning}, door opening \cite{gu2017deep}, manipulation \cite{levine2016end,mohtasib2021study}, and block insertion \cite{johannink2019residual}. Unfortunately, most results from applying DRL to robotics have been attained in simulation due to the large number of agent-environment interactions required and exploration challenges in the real world.

{\bf Combining Reinforcement Learning with Demonstrations}.  There is a growing interest in combining imitation learning with RL to mitigate the problems of using pure imitation learning or pure RL. 
Reinforcement learning from demonstration (RLfD) approaches concern scenarios in which the expert also receives rewards from the environment \cite{piot2014boosted}. Most of these approaches adopt value-based RL algorithms, which are off-policy in nature, as they are more data-efficient. For instance, DQfD \cite{hester2018deep} introduces RLfD into DQN \cite{mnih2015human} by adding demonstration data into a prioritised replay buffer \cite{schaul2016prioritized}. However, DQfD is limited to applications with discrete action spaces. Vecerık et al. \cite{vecerik2017leveraging} developed DDPGfD, which extends RLfD to continuous action domains. DDPGfD \cite{vecerik2017leveraging} is built upon DDPG \cite{lillicrap2016continuous} similarly to DQfD \cite{hester2018deep}. Both DQfD and DDPGfD suffer the problem of under-exploiting demonstration data, as explained by Kang et al. \cite{kang2018policy}. Nair et al. \cite{nair2018overcoming} extend the DDPGfD algorithm \cite{vecerik2017leveraging} by introducing HER \cite{andrychowicz2017hindsight} to it, and by adding a Q-filter to the loss. Similarly, Zuo et al. \cite{zuo2020efficient} introduced HER \cite{andrychowicz2017hindsight} to TD3 \cite{fujimoto2018addressing}. However, HER \cite{andrychowicz2017hindsight} assumes pre-knowledge of the final goal state of the task, which is sometimes not possible in robotics and limits its applications to real robots.

Some existing RLfD methods use on-policy RL to leverage the demonstration data. Kim et al. \cite{kim2013learning} and Piot et al. \cite{piot2014boosted} algorithms are based on policy iteration and use demonstration data to shape the value function. Those methods however need the value of the expert state-action pairs to be larger than the others with a margin. Thus, they suffer from performance decline when the demonstration data is not perfect. DAPG \cite{Rajeswaran-RSS-18} and POfD \cite{kang2018policy} have shown state-of-the-art results on a variety of tasks by combining the original RL loss with a behaviour cloning loss on the expert’s demonstrations. Consequently, the agent simultaneously learns its original objective and the behaviour of the expert using on-policy gradient methods \cite{Rajeswaran-RSS-18,kang2018policy}. DAPG bootstraps the policy using behavioural cloning and data augmentation to learn several complex manipulation tasks \cite{Rajeswaran-RSS-18}. DAPG is similar to our method in bootstrapping the policy using behavioural cloning. In contrast, our algorithm uses off-policy RL and novel  advantage weighting and early termination tricks.

Closely related to our method are the learning algorithms proposed by Nair et al. \cite{nair2020accelerating} (AWAC), Wang et al. \cite{wang2020critic} (CRR), and Peng et al. \cite{peng2018deepmimic} (DeepMimic). AWAC weights the policy optimisation objective with the estimated action advantage $A(s,a)$, and CRR is very similar to AWAC. Unlike AWAC and CRR, our algorithm calculates the agent advantage w.r.t the expert Q-values to help the agent outperform imperfect expert data. In addition, AWAC and CRR do not use early termination when fine-tuning the policy. Unlike DeepMimic, which uses task-specific predefined early termination conditions, our algorithm uses automatic early termination triggers based on the episode similarity to expert data.

Another class of related works perform offline reinforcement learning by training only from previously collected offline data from other policies \cite{levine2020offline,agarwal2020optimistic,fujimoto2019off,fu2020d4rl,wu2019behavior,siegel2020keep,kumar2020conservative,peng2019advantage,kumar2019stabilizing}. Unfortunately, offline RL methods require  large amounts of offline data and suffer from error accumulation with distribution shift when fine-tuning with online data. Thus, we are not including them in our study. 







\section{ALGORITHM}

This section describes our proposed advantage weighting with early termination actor-critic (AWET) algorithm. AWET can be applied to any off-policy actor-critic algorithm, and here we show how to apply it to TD3 \cite{fujimoto2018addressing}. AWET consists of two stages: (1) offline training using offline expert data, see section \ref{offline_traning}; and (2) fine-tuning using both offline data and online data collected by the agent, see section \ref{online_training}. 

\subsection{Offline Training Stage}
\label{offline_traning}
In this stage, we use supervised training on the offline expert data to train an ensemble of critics and then  train the policy. We start by calculating the state-action value $Q_E(s,a)$ for each state-action pair in every trajectory $\tau$ of the expert's replay buffer $D_E$ using Monte Carlo estimation, i.e.,

\begin{equation}
\label{eqn_1}
Q_E(s,a)=\mathbb{E}_{\tau \in D_E}\Big[\sum_{k=0}^{|\tau|-t-1} \gamma^k r_{t+k+1}\big|s_t,a_t \Big].
\end{equation}
We then optimise every critic $i$ in the ensemble of critics (for simplicity, we assume that we have only two critics) with the mean square error (MSE) and $L2$ regularisation objectives according to equation~\ref{eqn_2}. The $L2$ regularisation objective $\mathcal{L}_{L2}$ helps to prevent overfitting on the expert data $B_E$. Hence, the overall loss function of the $i$-th critic is therefore given by

\begin{equation}
\label{eqn_2}
\mathcal{L}_{\phi_i}=\frac{1}{\left|B_E\right|}\sum_{\left(s,a\right)\in B_E} \big\Vert Q_{\phi_i}\left(s,a\right)-Q_E\left(s,a\right)\big\Vert^2+\lambda_1 \mathcal{L}_{L2}.
\end{equation}




Finally, the trained critics are used to update the policy. The standard objective for updating the policy in TD3 \cite{fujimoto2018addressing} is to maximise the Q-value of the first critic $Q_{\theta_1}$. First, we extend this object to use the critic with the minimum Q-value: $\mathcal{L}_Q=\frac{1}{\left|B_E\right|}\sum_{s\in B_E} \min_{i=1,2}  Q_{\phi_i}\left(s,\mu_{\theta}\left(s\right)\right)$.
Second, we introduce an additional behavioral cloning loss computed on the offline expert data for training the policy \cite{Rajeswaran-RSS-18,nair2018overcoming}: 
\begin{equation}
\label{eqn_4}
\mathcal{L}_{BC}=\frac{1}{\left|B_E\right|}\sum_{(s, a)\in B_E}  \big\Vert\mu_{\theta}\left(s\right)- a \big\Vert^2,
\end{equation}
where $a$ represents the actions from the expert behavioural policy. We use this loss as an auxiliary loss by weighting it with the hyperparameter $C_l$, which aims to improve learning performance. The $\mathcal{L}_{BC}$ loss forces the policy to select actions that are, to some extent, similar to the expert actions. Thus, choosing the $C_l$ weight correctly is essential as we want the policy to outperform the expert, not only mimic the expert. Hence, $C_l$ performs a reasonable trade-off between the $\mathcal{L}_Q$ loss and the $\mathcal{L}_{BC}$ loss. Similarly to the critics, we also add a $L2$ regularisation objective $\mathcal{L}_{L2}$ with a weights decay factor $\lambda_2$ to prevent overfitting to the actor. The overall loss function of the actor is given by

\begin{equation}
\label{eqn_5}
\mathcal{L}_{\theta}=-\left(1-C_l\right)\mathcal{L}_Q + C_l\mathcal{L}_{BC} + \lambda_2  \mathcal{L}_{L2}.
\end{equation}


Note that we are maximising $\mathcal{L}_Q$ and minimising $\mathcal{L}_{BC}$. The offline training stage of AWET is listed in Algorithm \ref{awet_offline}.

\begin{algorithm}[t!]
\renewcommand{\thealgorithm}{1}
\caption{AWET: Offline Training Stage}
\label{awet_offline}
\begin{algorithmic}[1]
    \INPUT{initial policy parameters $\theta$, Q-function parameters $\phi_1$, $\phi_2$, expert's data replay buffer $D_E$}
    \State Calculate the Q-value for $D_E$ according to eq. (\ref{eqn_1}).
    
    \For {t=1 \textbf{to} \texttt{gradient}\_\texttt{steps}}
        \State Sample a batch $B_E=\{\left(s,a,r,s^{\prime},d\right)\}$ from $D_E$
        \State Update Q-functions according to eq. (\ref{eqn_2}).
    \EndFor

    \For {t=1 \textbf{to} \texttt{gradient}\_\texttt{steps}}
        \State Sample a batch $B_E=\{\left(s,a,r,s^{\prime},d\right)\}$ from $D_E$
        \State Update the policy according to eq. (\ref{eqn_5}).
    \EndFor
    \OUTPUT{policy parameters $\theta$, Q-function parameters $\phi_1$, $\phi_2$, expert's replay buffer $D_E$ with calculated Q-values}
\end{algorithmic}
\end{algorithm}

\subsection{Online Fine-Tuning Stage}
\label{online_training}

This stage fine-tunes the critics' ensemble and the policy using both the offline expert data and the online agent interactions with the environment (See Algorithm \ref{awet_online}). AWET thus uses two mini-batches for training: $B_E$ is sampled from the expert's replay buffer $D_E$, and $B_A$ is sampled from the agent's replay buffer $D_A$. We start by collecting rollouts using the current policy. For each rollout, we compute its similarity to the offline expert data, and discard any rollout that has a similarity less than a threshold value. Subsequently, we store any undiscarded rollouts in the agent's replay buffer and compute the Q-values for the agent's transitions which are finally used to estimate the agent's advantage. Finally, we update the Q-functions and the policy using carefully designed objectives as in Eqs.~\ref{eqn_11} and ~\ref{eqn_14}. 

\begin{algorithm}[t!]
\renewcommand{\thealgorithm}{2}
\caption{AWET: Online Fine-Tuning Stage}
\label{awet_online}
\begin{algorithmic}[1]
    \INPUT{offline trained policy parameters $\theta$, offline trained Q-function parameters $\phi_1$, $\phi_2$, expert's replay buffer $D_E$, empty agent's replay buffer $D_A$}
    \State Set target networks ${\theta^{\prime}} \gets \theta$, ${\phi^{\prime}}_1 \gets \phi_1$, ${\phi^{\prime}}_2 \gets \phi_2$
    \State Calculate $S_{th}$ as described in section \ref{dtw} 
    \Repeat 
        \While {not \texttt{done}}  \Comment{collect rollouts}
            \State Observe state $s$ and execute action $a$
            \[ 
                a=\mu_{\theta}(s)+\epsilon, \tag*{$\epsilon \sim \mathcal{N}(0,\sigma)$}
            \]

            \State Observe next state $s^{\prime}$, reward $r$, and done $d$
            \If {reached half of the episode ($\frac{max.steps}{2}$)}
                \If {$\min\limits_{\forall \tau \in D_E}DTW(rollout,\tau) > S_{th}$}
                    \State Terminate and discard the transitions
                \EndIf
            \EndIf
        \EndWhile
        
        \State Store undiscarded transitions $\{(s,a,r,s^{\prime},d)\}$ in $D_A$
        \State Reset environment state.
        \If {it's time to update} 
            \For {$j$ in range (however many updates)}
                \State Sample mini-batches: $B_E\sim D_E$, $B_A\sim D_A$
                \State Compute target actions for data in $B_A$
                \[ 
                a^{\prime}(s^{\prime})=\mu_{{\theta^{\prime}}}(s^{\prime})+\epsilon,  \tag*{ $\epsilon \sim clip(\mathcal{N}(0,\tilde{\sigma}),-c,c)$}
                \]
                
                \State Compute targets $y(r,s',d)$ for data in $B_A$ 
                \State Compute $A_A$ using eq. (\ref{eqn_7})
                \State Update the Q-functions according to eq. (\ref{eqn_11})
                
                \If{$j\mod$ \texttt{policy}\_\texttt{delay }$=0$}
                    \State Update the policy according to eq. (\ref{eqn_14})
                    \State Update target networks with
                    \[
                    {\phi^{\prime}}_i \leftarrow \rho {\phi^{\prime}}_i+\left(1-\rho \right) \phi_i,  \tag*{${\theta^{\prime}} \leftarrow \rho {\theta^{\prime}}+\left(1-\rho \right) \theta$}
                    \]
                    
                \EndIf
            \EndFor
        \EndIf
    \Until{convergence}
\end{algorithmic}
\end{algorithm}

\label{dtw}
\textbf{Early Termination.} We propose a novel automatic early termination technique based on policy rollout similarity to expert rollouts via a threshold using the offline data with Dynamic Time Warping (DTW) \cite{berndt1994using,muller2007dynamic}. DTW is a well-known technique to find an optimal alignment between two given (time-dependent) sequences under certain restrictions \cite{berndt1994using,muller2007dynamic}. We calculate DTW distances between each trajectory pairs from the expert. The DTW distance threshold, referred to as $S_{th}$, is the average of all computed DTW distances. Assuming that we have $M$ trajectories in the offline expert data, $S_{th}$ is computed according to 
\begin{equation}
\label{eqn_6}
S_{th}=\frac{1}{(M^2-M)/2}\sum_{i=0}^{M-2} \sum_{j=0}^{i} DTW(\tau_i,\tau_j).
\end{equation}
To explain this equation in more detail, 
first, we do not need to compute the DTW distance between a trajectory and itself as this value will always be zero, i.e, $DTW(\tau_{i},\tau_{i})=0$. Second, the DTW distance between two trajectories is symmetric, i.e., $DTW(\tau_{i},\tau_{j})=DTW(\tau_{j},\tau_{i})$. Thus, we need only to compute $(M^2-M)/2$ distances and then find their average to get the DTW distance threshold, $S_{th}$. 



In each policy rollout, after completing half of the rollout length, we compute its DTW distance, $S$, to all trajectories in the expert data, $D_E$, then take the minimum, i.e., $S=\min_{\forall \tau \in D_E}DTW(rollout,\tau)$. If this DTW distance is larger than $S_{th}$, we trigger early termination and discard this rollout data. Otherwise, we complete the rollout and store its data in $D_A$. So, if the policy was performing a rollout that is not similar to any trajectory in the expert data, due to exploration or due to any other reason, then this rollout is non-useful, and it may even lead the policy to exploit bad behaviours if used for training the policy. Thus, we terminate the non-useful rollout and discard its data. 

\begin{table*}[t!]
\vspace{4mm}
\caption{Experimental results comparing various baseline algorithms against AWET in four different tasks.}
\vspace{-6mm}
\begin{center}
\begin{tabular}{|c|c|c|c|c|c|c|c|c|}
\hline
      & \multicolumn{2}{c|}{\textbf{Pendulum}} & \multicolumn{2}{c|}{\textbf{Reacher}} & \multicolumn{2}{c|}{\textbf{Pusher}} & \multicolumn{2}{c|}{\textbf{Fetch (Reach)}} \\ \hline
      & \multicolumn{2}{c|}{\includegraphics[scale=0.138]{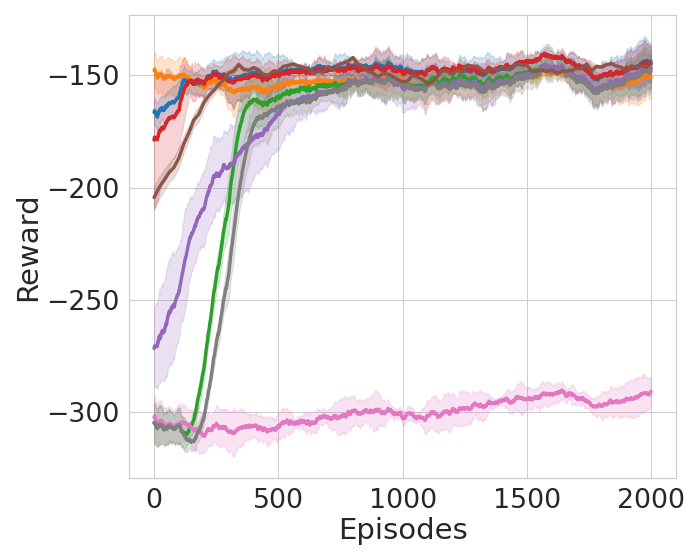}} & \multicolumn{2}{c|}{\includegraphics[scale=0.138]{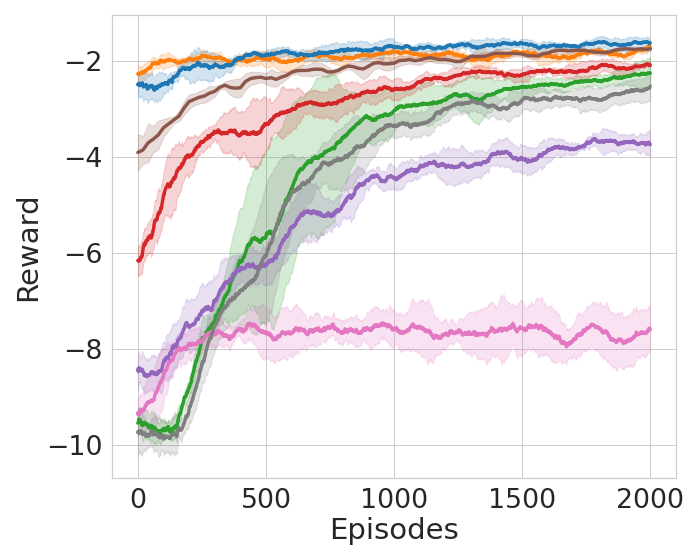}}  &  \multicolumn{2}{c|}{\includegraphics[scale=0.138]{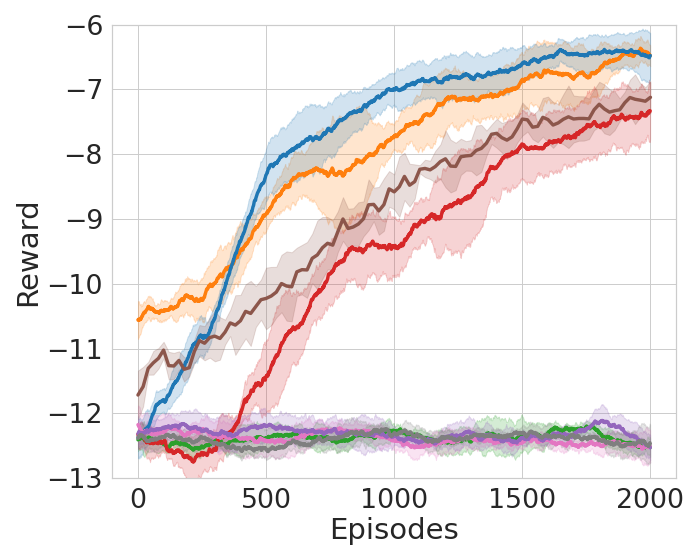}} &  \multicolumn{2}{c|}{\includegraphics[scale=0.138]{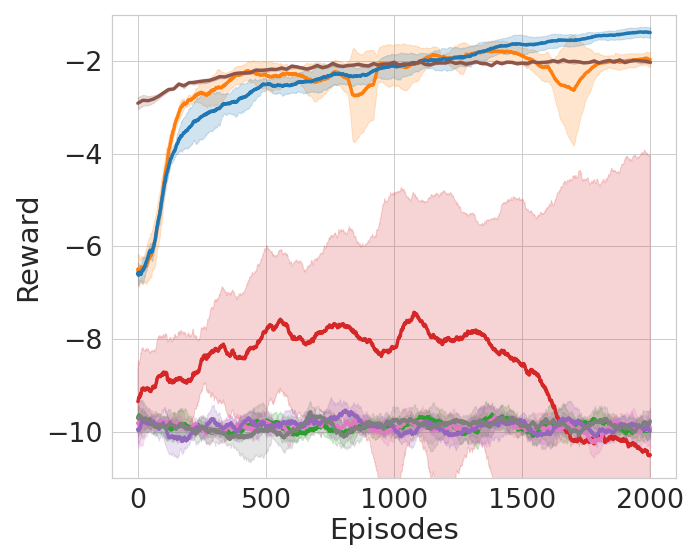}}   \\ \hline
      & \multicolumn{8}{c|}{\includegraphics[scale=0.80]{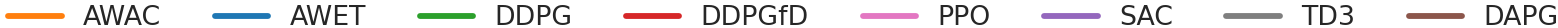}} \\ \hline
                         & Success Rate               & Rewards              & Success Rate              & Rewards              & Success Rate              & Rewards             & Success Rate                 & Rewards                 \\ \hline
    \textbf{DDPG}        & \scriptsize $88.6\%\pm24\%$ & \scriptsize $-145.8\pm11.5$ & \scriptsize $91.8\%\pm5.2\%$ & \scriptsize $-3.4\pm0.5$ & \scriptsize $0\%\pm0\%$ & \scriptsize $-24.9\pm0.4$ & \scriptsize $0\%\pm0\%$ & \scriptsize $-11.4\pm1.6$ \\ \hline
    \textbf{TD3}         & \scriptsize $78.8\%\pm44\%$ & \scriptsize $-151.5\pm21.2$ & \scriptsize $82.8\%\pm10.5\%$ & \scriptsize $-4.3\pm0.7$ & \scriptsize $0.2\%\pm0\%$ & \scriptsize $-24.6\pm0.5$ & \scriptsize $0.0\%\pm0.0\%$ & \scriptsize $-10.8\pm0.9$ \\ \hline
    \textbf{SAC}         & \scriptsize $79.4\%\pm44\%$ & \scriptsize $-151.6\pm10$ & \scriptsize $32.4\%\pm4.4\%$ & \scriptsize $-7.3\pm0.7$ & \scriptsize $0\%\pm0\%$ & \scriptsize $-25\pm1.5$ & \scriptsize $0.0\%\pm0.0\%$ & \scriptsize $-11.3\pm1.1$ \\ \hline
    \textbf{PPO}         & \scriptsize $0.6\%\pm0.8\%$ & \scriptsize $-589.7\pm13.6$ & \scriptsize $7.6\%\pm3.5\%$ & \scriptsize $-13.5\pm0.6$ & \scriptsize $0\%\pm0\%$ & \scriptsize $-24.7\pm0.5$ & \scriptsize $0.0\%\pm0.0\%$ & \scriptsize $-12.0\pm1.5$ \\ \hline
    \textbf{DDPGfD}      & \scriptsize $92.4\%\pm11.4\%$ & \scriptsize $-145.1\pm12.1$ & \scriptsize $92.5\%\pm3.3\%$ & \scriptsize $-2.8\pm0.8$ & \scriptsize $47.8\%\pm13.4\%$ & \scriptsize $-14.0\pm0.9$ & \scriptsize $5.5\%\pm40.2\%$ & \scriptsize $-9.9\pm2.3$ \\ \hline
    \textbf{DAPG}        & \scriptsize $96.0\%\pm7.0\%$ & \scriptsize $-140.3\pm15.7$ & \scriptsize $99.4\%\pm0.9\%$ & \scriptsize $-2.2\pm0.4$ & \scriptsize $51.4\%\pm12.6\%$ & \scriptsize $-11.8\pm0.8$ & \scriptsize $66.8\%\pm14.5\%$ & \scriptsize $-2.9\pm0.7$  \\ \hline
    \textbf{AWAC}        & \scriptsize $99.2\%\pm0.8\%$ & \scriptsize $-135.9\pm8.8$ & \scriptsize $100\%\pm0\%$ & \scriptsize $-2.2\pm0.2$ & \scriptsize $71.6\%\pm10.5\%$ & \scriptsize $-11.2\pm1.3$ & \scriptsize $65.6\%\pm18.2\%$ & \scriptsize $-3.1\pm0.9$  \\ \hline
    \textbf{AWET} (ours) & \scriptsize {\cellcolor{gray!20}$100\%\pm0\%$} & \scriptsize {\cellcolor{gray!20}$-130.8\pm10$} & {\cellcolor{gray!20}$100\%\pm0\%$} & \scriptsize {\cellcolor{gray!20}$-1.9\pm0.3$} & \scriptsize {\cellcolor{gray!20}$85\%\pm7\%$} & \scriptsize {\cellcolor{gray!20}$-9.7\pm0.6$} & \scriptsize {\cellcolor{gray!20}$88\%\pm11\%$} & \scriptsize {\cellcolor{gray!20}$-2.0\pm0.4$} \\ \hline
\end{tabular}
\label{results_tab1}
\end{center}
\vspace{-4mm}
\end{table*}

\textbf{Advantage Weighting.} Before updating the critics and the actor, AWET calculates the agent's advantage $A_A$ to weigh the critics' losses. $A_A$ is a single weight for the whole batch, and it is not possible to compute this weight per sample because the states in $B_E$ might be different from the states in $B_A$. $A_A$ measures how much better the agent state-action pairs are compared to the expert pairs in the current training mini-batches, which forces the agent to outperform offline expert data. This is novel and differ from the well-known action advantage value $A(s,a)$ that has been used by different RL algorithms. Then, AWET weights the agent data losses in the critics objective by ($A_A$) and the expert data losses by ($1-A_A$). $A_A$ is calculated according to

\begin{equation}
\label{eqn_7}
A_A=\frac{\frac{1}{\left|B_A\right|}\sum_{\left(s,a\right)\in B_A} Q_{\phi_i}\left(s,a\right)} {\frac{1}{\left|B_A\right|}\sum_{\left(s,a\right)\in B_A} Q_{\phi_i}\left(s,a\right)+\frac{1}{\left|B_E\right|}\sum_{\left(s,a\right)\in B_E} Q_{\phi_i}\left(s,a\right)}.
\end{equation}

Here, we should note that AWET calculates the average Q-values using the $i$-th critic that has the minimum Q-values. Also, we assume that the rewards, both for the expert and the agent datasets, are normalized and always have the same sign---either positive or negative.

\textbf{Training Objectives.} The first objective in the critics' update is the original TD3 objective on $B_A$  calculated as 

\begin{equation}
\label{eqn_8}
\mathcal{L}_{B_A}=\frac{1}{\left|B_A\right|}\sum_{\left(s,a,r,s^{\prime},d\right)\in B_A} \big\Vert Q_{\phi_i}\left(s,a\right)-y\left(r,s^{\prime},d\right)\big\Vert^2,
\end{equation}
where $y\left(r,s^{\prime},d\right)$ is the target Q-values using the target Q-nets ${\phi^{\prime}}_i$, as $y(r,s^{\prime},d)=r+\gamma(1-d)\min_{i=1,2}Q_{{\phi^{\prime}}_i}(s^{\prime},a^{\prime}(s^{\prime}))$ and $d$ is the done signal to indicate the end of the trajectory.


It is known that the critics may overestimate the Q-values when having out-of-distribution actions as expanded by Kumar et al. \cite{kumar2019stabilizing}. There are different ways to deal with this problem, and we choose clipping the loss function, which is a simple solution to the problem. AWET uses an auxiliary MSE objective on the expert's Monte-Carlo returns in data set $B_E$. This objective helps in preventing any drastic shift away from the expert distribution. To the extent of our knowledge, the mixing loss on Monte-Carlo returns and TD loss is novel. This auxiliary loss is calculated according to

\begin{equation}
\label{eqn_10}
\mathcal{L}_{B_E}=\frac{1}{\left|B_E\right|}\sum_{\left(s,a,r,s^{\prime},d\right)\in B_E} \big\Vert Q_{\phi_i}\left(s,a\right)-Q_E\left(s,a\right)\big\Vert^2.
\end{equation}

Thus, the overall loss function of the $i$-th critic in the online fine-tuning stage is given by


\begin{equation}
\label{eqn_11}
\mathcal{L}_{\phi_i}=A_A clip(\mathcal{L}_{B_A},-C_{clip},C_{clip})+(1-A_A) \mathcal{L}_{B_E}.
\end{equation}

Based on the above, we can use the fine-tuned critics in the actor objective. AWET applies the original TD3 actor objective to the data in both mini-batches $B_A$ and $B_E$ for policy learning by maximising the Q-values. These losses are computed according to $\mathcal{L}_{Q_A}=\frac{1}{\left|B_A\right|}\sum_{s\in B_A} \min_{i=1,2} Q_{\phi_i}\left(s,\mu_{\theta}\left(s\right)\right)$ and $\mathcal{L}_{Q_E}=\frac{1}{\left|B_E\right|}\sum_{s\in B_E} \min_{i=1,2} Q_{\phi_i}\left(s,\mu_{\theta}\left(s\right)\right)
$.

%

Similarly to the critics' objective, AWET uses an auxiliary MSE objective on the expert's data $B_E$ to do not deviate too much from the expert's policy and to eliminate or reduce rapid shifts away from the expert distribution. The actor auxiliary loss is calculated using eq. (\ref{eqn_4}). 
The overall loss function of the actor in the online stage is defined as


\begin{equation}
\label{eqn_14}
\mathcal{L}_{\theta}=-\left(1-C_l\right)\mathcal{L}_{Q_E} + C_l\mathcal{L}_{BC} - \mathcal{L}_{Q_A}.
\end{equation}

Recall that in this objective we are maximising $\mathcal{L}_{Q_A}$ and  $\mathcal{L}_{Q_E}$ while  minimising $\mathcal{L}_{BC}$.

\section{EXPERIMENTS}
\label{experiments}

\subsection{Experimental Setup}

\begin{figure}[t!]
\vspace{2mm}
    \centerline{\includegraphics[width=\columnwidth]{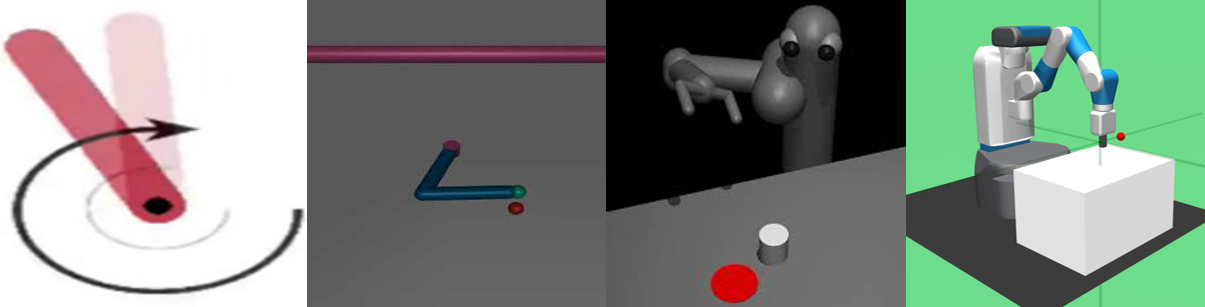}}
    \caption{Visualisation of tasks: Pendulum, Reacher, Pusher, Fetch (Reach).}
    \label{fig_envs}
\end{figure}

\textbf{Environments.} We evaluate our proposed method on four  robotic manipulation tasks using the Mujoco simulator \cite{todorov2012mujoco} within OpenAI Gym \cite{brockman2016openai}: Pendulum, Reacher, Pusher, and FetchReach. These tasks, shown in Fig.~\ref{fig_envs}, have different complexity, ranging from 1 Degree-of-Freedom (DoF) in the Pendulum task up to 7 DoF in the Pusher and Fetch tasks.  

\textbf{Demonstration Data Collection.} To generate demonstrations, we used a pre-trained expert policy. For each task, we generated 100 successful trajectories of 50 timesteps long. We used different portions of these demonstrations in our experiments, as explained in the next sections. Each demonstration is a whole trajectory and each timestep in this trajectory can be expressed as a tuple $<s,a,s^{\prime},r,d>$. For simplicity, we used oracle dense rewards calculated using the physics simulator of the task.


\textbf{Training Details.} All agents were trained for 1000 gradient steps in the offline stage and for 2000 episodes (100,000 timesteps) in the online fine-tuning stage---using the Adam optimiser \cite{kingma2014adam} with learning rate $10^{-3}$. This training budget is very small compared to the timesteps needed to train standard RL. In our experiments, the discount factor $\gamma$ is 0.98. The policy and the Q-functions are feedforward neural networks with ReLU activation. All neural networks have two layers of sizes [400, 300]. The final layer of the policy uses $\textrm{tanh}$ activation, and the output is scaled to the task's action ranges. In the offline training stage, we used L2 regularisation, while no regularisation is used in the online fine-tuning stage. 


\textbf{Overview of Experiments.} We compare our algorithm to previous work in Section~\ref{exp1}, and perform ablations of our proposed method in Section~\ref{exp2}.

\subsection{Comparison with Prior Work}
\label{exp1}

We compared AWET against the  state-of-the-art RL algorithms PPO \cite{schulman2017proximal}, SAC \cite{haarnoja2018soft}, TD3 \cite{fujimoto2018addressing}, and DDPG \cite{lillicrap2016continuous}. These algorithms are trained using only online data collected during training. We also compare AWET to some of the state-of-the-art RLfD algorithms including AWAC \cite{nair2020accelerating}, DAPG \cite{Rajeswaran-RSS-18}, and DDPGfD \cite{vecerik2017leveraging}---seven baselines in total. We perform this comparative study in the four robotic tasks shown in Fig.~\ref{fig_envs}.
Table~\ref{results_tab1} presents the learning curves of our comparative study of those algorithms in the four different tasks. In addition,  Table~\ref{results_tab1} presents the success rate of each algorithm in the different tasks averaged across 10 runs. For AWET and all other RLfD algorithms used in this study, we used the same expert data (100 expert demonstration episodes/trajectories) for training the agents, and for each task.

The results in Table~\ref{results_tab1} show that the RLfD algorithms outperform the RL algorithms in terms of the success rates, and they converged faster. This shows that the offline expert data is indeed helpful to speed up the online learning process by directing the policy to the right behaviours. Overall, the results of the RLfD algorithms in Table~\ref{results_tab1} show that AWET was able to achieve competitive performance in comparison to the state-of-the-art RLfD algorithms AWAC \cite{nair2020accelerating}, DAPG \cite{Rajeswaran-RSS-18}, and DDPGfD \cite{vecerik2017leveraging} in the four tasks. Although the Fetch (Reach) task proved to be the most challenging case,  AWET was able to outperform all baselines. 
This suggests that AWET's tricks are key for effective performance.

A statistical analysis using the Wilcoxon Signed-Rank Test (paired) \cite{wilcoxon1992individual} on the results of Table~\ref{results_tab1} revealed that the success rate differences between AWET against AWAC, DAPG and DDPGfD are significant at $p=0.031$, $p=0.0078$, and $p=8e-5$, respectively. This is promising and supports that AWET has outperformed AWAC, DAPG, and DDPGfD.

\subsection{Ablation Experiments}
\label{exp2}

Our ablation study here helps to analyse the different components of AWET on the most challenging robotic task in our experiments (Pusher) with 10 runs in each test case. We start by studying the effect of the number of expert demonstrations. In this experiment, we train AWET with different amounts of offline data for 1000 offline gradient steps and a batch size of 100. The results in Table~\ref{results_tab2}, show that increasing the number of expert training demonstrations---from 20 demos to 100 demos---leads to increased  performance ($50.7\%\pm14.6\%$ and $85\%\pm7\%$ success rates, respectively) and faster convergence. Although not surprisingly (as more expert demonstrations mean more information about the task at hand),  more training data may require more gradient steps to leverage all information from it.


\begin{table}[t!]
\vspace{2.2mm}
\caption{Results of learning agents using different amounts of demonstrations in the Pusher task.}
\begin{center}
\begin{tabular}{|c|c|c|c|}
\hline
\multirow{12}{*}{\includegraphics[scale=0.167]{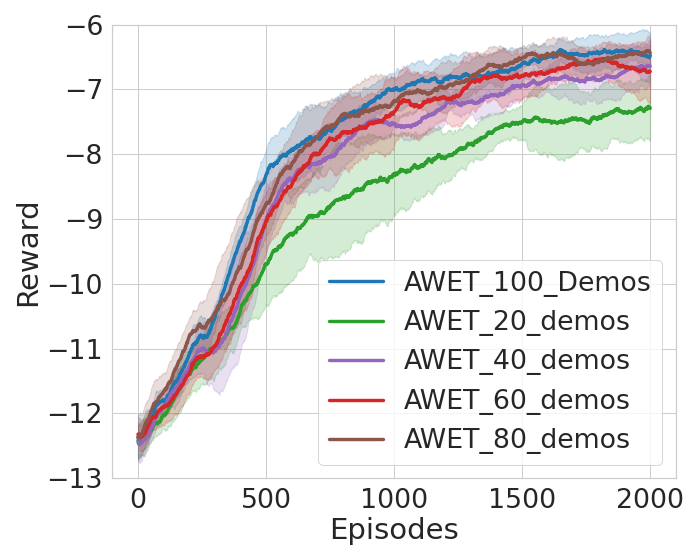}}
                  & \multirow{2}{*}{ \textbf{\# Demos}}                    & \multirow{2}{*}{\textbf{Success Rate}} \\ 
                  &  &      \\ \cline{2-3} 
                  & \multirow{2}{*}{20}             & \multirow{2}{*}{$50.7\%\pm14.6\%$} \\  & & \\  
                  & \multirow{2}{*}{40}             & \multirow{2}{*}{$64.3\%\pm11.3\%$} \\  & & \\  
                  & \multirow{2}{*}{60}             & \multirow{2}{*}{$68.1\%\pm9.9\%$}  \\  & & \\  
                  & \multirow{2}{*}{80}             & \multirow{2}{*}{$69.6\%\pm7.8\%$}  \\  & & \\  
                  & \multirow{2}{*}{100}            & \multirow{2}{*}{$85\%\pm7\%$} \\  & & \\  \hline
\end{tabular}
\label{results_tab2}
\end{center}
\vspace{-5mm}
\end{table}

Moreover, we studied the effect of the agent's advantage $A_A$ on the learned policy. We have tested different ways to calculate $A_A$: (1) applying Eq.~\ref{eqn_7} to the critic losses, (2) applying the $softmax$ function to the Q-values, and (3) using Eq.~\ref{eqn_7}---as shown in Table~\ref{results_tab3}. The results in this table show significant improvement 
when $A_A$ is calculated using the Q-values according to Eq.~\ref{eqn_7} compared to the case when $A_A$ is not used at all---the success rate is $85\%\pm7\%$ and $70.8\%\pm11.7\%$, respectively. When $A_A$ is calculated using the critic losses, we found that it causes a drop in the agent's performance (from $70.8\%\pm11.7\%$ without using $A_A$ to $64.9\%\pm9.3\%$ when using $A_A$ calculated using the critic losses). That is because when $A_A$ is calculated using the critic losses, it loses important information on how better the agent's performance is compared to the expert. Similarly, using softmaxed Q-values to calculate $A_A$ showed the worst results due to its sparse behaviour, causing high variance.

\begin{table}[t]
\vspace{2mm}
\caption{Ablation results of Advantage Weighting ($A_A$).}
\vspace{-1mm}
\begin{center}
\begin{tabular}{|c|c|c|}
\hline
\multirow{10}{*}{\includegraphics[scale=0.160]{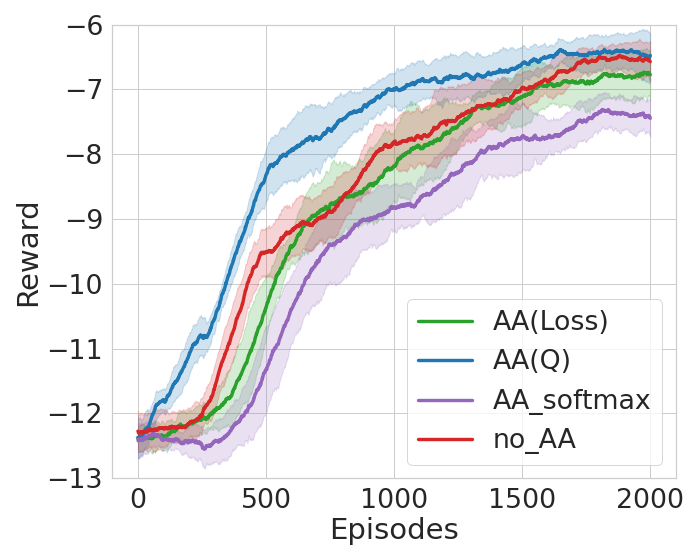}}
                  &   \multirow{2}{*}{\textbf{$A_A$}} & \multirow{2}{*}{\textbf{Success Rate}} \\   
                  &           &            \\ \cline{2-3} 
                  & \multirow{2}{*}{no $A_A$}                             & \multirow{2}{*}{$70.8\%\pm11.7\%$}   \\  & & \\  
                  & \multirow{2}{*}{$\frac{\mathcal{L}_{B_A}}{\mathcal{L}_{B_A}+\mathcal{L}_{B_E}}$}    & \multirow{2}{*}{$64.9\%\pm\ 9.3\%$}    \\  & & \\  
                  & \multirow{2}{*}{$\frac{e^{Q_A}}{e^{Q_A}+e^{Q_E}}$}    & \multirow{2}{*}{$55.4\%\pm11.0\%$}     \\  & & \\  
                  & \multirow{2}{*}{Eq. \ref{eqn_7}}                      & \multirow{2}{*}{$85\%\pm7\%$}   \\  & & \\  \hline
\end{tabular}
\label{results_tab3}
\end{center}
\vspace{-4mm}
\end{table}

\begin{figure}[t!]
\vspace{2mm}
    \begin{center}
        \begin{tabular}{cc}
        \includegraphics[scale=0.165]{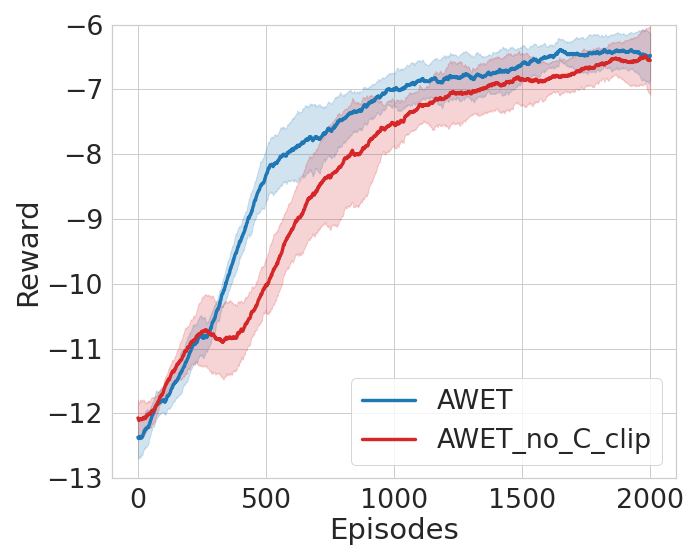} & \hspace{-10pt} \includegraphics[scale=0.165]{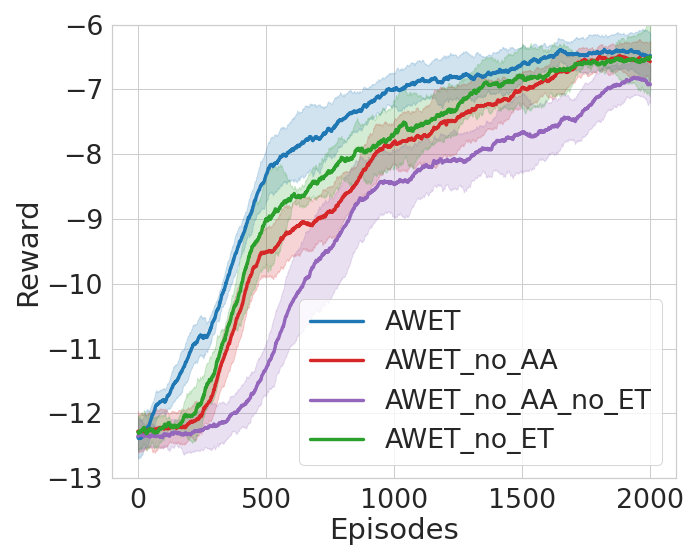}  \\
        \footnotesize{(A)}  & \footnotesize {(B)} \\
        \end{tabular}
    \end{center}
    \caption{(A) The effect of $C_{clip}$. (B) The effect of early termination.}
    \label{fig_results}
\end{figure}

Also, we have tested the effect of the critic loss clipping $C_{clip}$. When AWET uses $C_{clip}=0.5$, the success rate was $85\%\pm7\%$. However, without loss clipping, the success rate dropped to $70.2\%\pm14.1\%$ and with slower convergence, see Fig.~\ref{fig_results}(A). So, $C_{clip}$ helps in preventing the Q-functions from dramatically drifting from the expert distribution.

We also studied the effects of early termination. Fig.~\ref{fig_results}(B) shows learning curves of training different versions of AWET in the Pusher task. The full version of AWET that uses both agent's advantage weighting and automatic early termination achieved the best results ($93.25\%\pm4.5\%$ an average success rate) across tasks, see Table~\ref{results_tab4}. However, AWET with no early termination only achieved an average success rate of $77.3\%\pm11.3\%$ across tasks. That is because the agent trajectories that are not similar to the expert trajectories are less informative and even misleading in the learning process. 
Moreover, AWET with no $A_A$ has achieved an average success rate of $82.5\%\pm9.2\%$ across tasks. The worst-case scenario is when both agent's advantage weighting and automatic early termination are not used ($72.9\%\pm8.9\%$ average success rate across tasks). This proves that both the agent's advantage weighting and the early termination tricks help in improving the performance of AWET significantly.

Furthermore, we applied AWET to DDPG and SAC to study its effects. In the Pusher and Fetch (Reach) tasks, pure TD3, DDPG, and SAC have not shown any successful learning (almost $0\%$ success rate). 
In contrast, applying AWET to these algorithms showed significant improvements, as shown in Table~\ref{results_tab5}. We carried out a statistical analysis using the Wilcoxon Signed-Rank Test (paired) \cite{wilcoxon1992individual} to determine a ranking of algorithms according to their average success rates. Whilst the performance difference between AWET$^{(TD3)}$ and AWET$^{(DDPG)}$ is not significant ($p=0.398$), the differences between AWET$^{(TD3)}$ Vs. AWET$^{(SAC)}$ and AWET$^{(DDPG)}$ Vs. AWET$^{(SAC)}$ are significant at $p=0.001$ and $p=0.0034$, respectively. These results suggest that AWET is more effective with  deterministic off-policy algorithms than stochastic off-policy ones.

\begin{table}[t]
\caption{Ablation results per task over 10 runs of 2000 episodes each.}
\begin{center}
\vspace{-1mm}
\begin{tabular}{|l|l|c|c|}
\hline
\textbf{Task} & \textbf{Agent}       & \textbf{Success Rate} & \textbf{Rewards}  \\ \hline \hline
\multirow{4}{*}{Pendulum} &
AWET                 & \cellcolor{gray!20}$100.0\%\pm0\%$ & \cellcolor{gray!20}$-130.8\pm10$  \\ 
 & AWET\_no\_AA         & $99.7\%\pm0.9\%$     & $-146.3\pm12$ \\ 
& AWET\_no\_ET         & $100.0\%\pm0\%$       & $-143.5\pm9$ \\ 
& AWET\_no\_AA\_no\_ET & $99.7\%\pm0.9\%$       & $-146.3\pm13$ \\ \hline
\multirow{4}{*}{Reacher} &
AWET                 & \cellcolor{gray!20}$100\%\pm0\%$ & \cellcolor{gray!20}$-1.9\pm0.3$ \\ 
 & AWET\_no\_AA         & $96.9\%\pm4\%$     & $-2.1\pm0.3$ \\ 
& AWET\_no\_ET         & $97.5\%\pm4\%$       & $-2.2\pm0.2$ \\ 
& AWET\_no\_AA\_no\_ET & $99.0\%\pm1\%$       & $-2.1\pm0.1$ \\ \hline
\multirow{4}{*}{Pusher} &
AWET                 & \cellcolor{gray!20}$85\%\pm7\%$     & \cellcolor{gray!20}$-9.7\pm0.6$ \\ 
 & AWET\_no\_AA         & $72.1\%\pm11.9\%$     & $-10.7\pm0.8$ \\ 
& AWET\_no\_ET         & $64.2\%\pm10.0\%$       & $-11.4\pm1.0$ \\ 
& AWET\_no\_AA\_no\_ET & $45.0\%\pm14.3\%$       & $-13.8\pm1.9$ \\ \hline
\multirow{4}{*}{Fetch} &
AWET                 & \cellcolor{gray!20}$88\%\pm11\%$ & \cellcolor{gray!20}$-2.0\pm0.4$ \\ 
 & AWET\_no\_AA         & $61.2\%\pm20.1\%$     & $-2.3\pm0.8$ \\ 
& AWET\_no\_ET         & $47.8\%\pm30.5\%$       & $-2.4\pm0.6$ \\ 
& AWET\_no\_AA\_no\_ET & $48.0\%\pm19.6\%$       & $-2.7\pm1.0$ \\ \hline
\end{tabular}
\label{results_tab4}
\end{center}
\end{table}


\begin{table}[t!]
\caption{Results of AWET applied to SOTA algs. on challenging tasks.}
\begin{center}
\scriptsize
\begin{tabular}{|l|c|c|c|c|}
\hline
 \multirow{2}{*}{\textbf{Agent}}     & \multicolumn{2}{c|}{\textbf{Pusher}} & \multicolumn{2}{c|}{\textbf{Fetch (Reach)}} \\ \cline{2-5} 
                            & Success Rate   & Rewards        & Success Rate   & Rewards       \\ \hline \hline
    \textbf{DDPG}           & $0.0\%\pm0\%$  & $-24.9\pm0.4$  & $0.0\%\pm0\%$  & $-11.4\pm1.6$ \\ \hline
    \textbf{TD3}            & $0.2\%\pm0\%$  & $-24.6\pm0.5$  & $0.0\%\pm0\%$  & $-10.8\pm0.9$ \\ \hline
    \textbf{SAC}            & $0.0\%\pm0\%$  & $-25.0\pm1.5$  & $0.0\%\pm0\%$  & $-11.3\pm1.1$ \\ \hline \hline
    \textbf{AWET}$^{(DDPG)}$& $82\%\pm9\%$   & $-10\pm0.6$    & $84\%\pm14\%$  & $-2.1\pm0.3$  \\ \hline
    \textbf{AWET}$^{(TD3)}$ & $85\%\pm7\%$   & $-9.7\pm0.6$   & $88\%\pm11\%$  & $-2.0\pm0.4$  \\ \hline
    \textbf{AWET}$^{(SAC)}$ & $80\%\pm10\%$  & $-10.3\pm0.7$  & $80\%\pm15\%$  & $-2.2\pm0.6$  \\ \hline
\end{tabular}
\label{results_tab5}
\end{center}
\vspace{-4mm}
\end{table}

Lastly and to study the computational expenses of AWET, we report the time needed to train the DDPG, TD3, and SAC agents with and without AWET, as shown in Table~\ref{times_table}. It is tempting to assume that AWET is computationally more expensive and need a longer training times than when AWET is not used. That is due to the additional tricks that AWET introduces. On average, online training for 2000 episodes with AWET requires an additional $\approx$600 sec of training time compared to training for 2000 episodes without AWET\footnote{PC specs: \textbf{CPU}: Intel i7-6950 @ 3.00GHz, 10 cores. \textbf{RAM}: 32GB. \textbf{GPU}: NVIDIA TITAN X 12GB.}. In addition, AWET requires about 150 sec for 1000 gradient steps in the offline training stage. However, this cost comes with a huge benefit in terms of success rate and convergence speed. If one wants to train DDPG, TD3, and SAC agents without AWET and achieve the same success rate as AWET, they need to be trained for much more episodes. This means longer training times as illustrated in Table~\ref{times_table}. Even with the computational costs that AWET is accompanied with, it is still much more efficient than training RL agents purely from scratch---especially for complex tasks.

\begin{table}[t!]
\vspace{3mm}
\caption{Training time, in seconds, with and without AWET}
\begin{center}
\vspace{-2mm}
\begin{tabular}{|c|c|c|c|c|}
\hline

\multirow{2}{*}{Agent} &  \multicolumn{4}{c|}{Task} \\ \cline{2-5}
    & Pendulum & Reacher & Pusher & Fetch \\ \hline \hline
    
    \multicolumn{5}{|c|}{Training time for 2000 episodes without AWET} \\ \hline
    \textbf{DDPG}               & $396\pm2$ & $459\pm8$ & $516\pm6$ & $1221\pm8$ \\ \hline
    \textbf{TD3}                & $392\pm3$ & $441\pm10$ & $490\pm11$ & $1149\pm9$ \\ \hline
    \textbf{SAC}                & $806\pm5$ & $900\pm14$ & $988\pm12$ & $2136\pm32$ \\ \hline \hline
    
    \multicolumn{5}{|c|}{Training time for 2000 episodes with AWET} \\ \hline
    {\scriptsize\textbf{AWET}$^{(DDPG)}$}    & $998\pm26$ & $1039\pm18$ & $1154\pm32$ & $1972\pm47$ \\ \hline
    {\scriptsize\textbf{AWET}$^{(TD3)}$}     & $977\pm2$ & $1051\pm17$ & $1181\pm1$ & $2003\pm51$ \\ \hline
    {\scriptsize\textbf{AWET}$^{(SAC)}$}     & $1299\pm61$ & $1491\pm42$ & $1552\pm72$ & $2854\pm74$ \\ \hline \hline
    
    \multicolumn{5}{|p{0.95\columnwidth}|}{AWET was trained over 2000 episodes. To reach AWET's success rate, DDPG/TD3/SAC trained over 4000, 25000, 25000, and 75000 episodes in the Pendulum, Reacher, Pusher, and Fetch (Reach) tasks, respectively.} \\ \hline
    \textbf{DDPG}               & $801\pm2$ & $5752\pm13$ & $6500\pm9$ & $45723\pm71$ \\ \hline
    \textbf{TD3}                & $794\pm5$ & $5533\pm12$ & $6116\pm22$ & $43109\pm43$ \\ \hline
    \textbf{SAC}                & $1587\pm5$ & $11216\pm20$ & $12365\pm49$ & $80056\pm76$ \\ \hline
\end{tabular}
\label{times_table}
\end{center}
\vspace{-6mm}
\end{table}

\section{CONCLUDING REMARKS}

Motivated by the challenges faced by existing methods for learning robotic tasks, this paper  presents a novel actor-critic learning algorithm termed `AWET'. It uses offline expert data to bootstrap the agent's behaviour using an offline training stage, which in turn is fine-tuned by the actor-critic agent using an online training stage. AWET makes use of two novel techniques: (i) the actor losses are weighted towards the high Q-values, and (ii) the policy rollouts that are not similar to the expert trajectories are automatically terminated and discarded. AWET demonstrated superior (if not comparable) performance to state-of-the-art learning algorithms.


\section{ACKNOWLEDGMENT}

Work funded by the British Council HESPAL scholarship and the School of Computer Science at University of Lincoln.
\bibliographystyle{IEEEtran}
\bibliography{mybib}

\end{document}